# Robotics Vision-based Heuristic Reasoning for Underwater Target Tracking and Navigation

**Chua Kia & Mohd Rizal Arshad**
Underwater Robotics Research Group (URRG),
School of Electrical and Electronics Engineering, Universiti Sains Malaysia,
Penang, Malaysia,
rizal@eng.usm.my

*Abstract:* This paper presents a robotics vision-based heuristic reasoning system for underwater target tracking and navigation. This system is introduced to improve the level of automation of underwater Remote Operated Vehicles (ROVs) operations. A prototype which combines computer vision with an underwater robotics system is successfully designed and developed to perform target tracking and intelligent navigation. This study focuses on developing image processing algorithms and fuzzy inference system for the analysis of the terrain. The vision system developed is capable of interpreting underwater scene by extracting subjective uncertainties of the object of interest. Subjective uncertainties are further processed as multiple inputs of a fuzzy inference system that is capable of making crisp decisions concerning where to navigate. The important part of the image analysis is morphological filtering. The applications focus on binary images with the extension of gray-level concepts. An open-loop fuzzy control system is developed for classifying the traverse of terrain. The great achievement is the system's capability to recognize and perform target tracking of the object of interest (pipeline) in perspective view based on perceived condition. The effectiveness of this approach is demonstrated by computer and prototype simulations. This work is originated from the desire to develop robotics vision system with the ability to mimic the human expert's judgement and reasoning when maneuvering ROV in the traverse of the underwater terrain.

## 1. Introduction

Most of underwater pipeline tracing operations are performed by remote operated vehicles (ROVs) driven by human operators. These tasks are often requiring continue attention and knowledge/experience of human operators to maneuver the robot (Foresti. G.L. and Gentili.,2000). In these operations, human operators does not require an exact measurement from the visual feedback, but based on the reasoning.

For these reasons, it is desirable to develop robotics vision system with the ability to mimic the human mind (human expert's judgement of the terrain traverse) as a translation of human solution. In this way, human operators can be reasonably confident that decisions made by the navigation system are sound to ensure safety and mission completion. To achieve such confidence, the system can be trained by expert (Howard. A. et al, 2001). In order to enable robots to make autonomous decision that guide them through the most traversable regions of the terrain, fuzzy logic techniques can be developed for classifying traverse using computer vision-based reasoning. Computing with words is highly recommended either when the available information is too imprecise to use numbers or when there is a tolerance for imprecision which can be exploited to get tractability and a suitable interface with the real world (Zadeh. L, 1999).

Current position based navigation techniques cannot be used in object tracking because the measurement of the position of the interested object is impossible due to its unknown behavior (Yang Fan and Balasuriya, A, 2000).

The current methods available to realize target tracking and navigation of an AUV used optical, acoustic and laser sensors. These methods have problems mainly in terms of complicated processing requirement and hardware space limitation on AUVs (Yang Fan and Balasuriya, A, 2000). Other relevant research consists of neural-network based classifier of the terrain can be found in (Foresti. G.L. and Gentili.,2000) and (Foresti, G. L. and Gentili, 2002).Also, existing method using Hough transform and Kalman filtering for image enhancement has also been very popular (Tascini, G. et al, 1996), (Crovatot, D. et al, 2000), (El-Hawary, F. and Yuyang, Jing, 1993) , (Fairweather, A. J. R. et al ,1997) and (El-Hawary, F. and Yuyang, Jing, 1995).

## 2. Research Approach

Visible features of underwater structure enable humans to distinguish underwater pipeline from seabed, and to see individual parts of pipeline. A machine vision and image processing system capable of extracting and classifying these features is used to initiate target tracking and navigation of an AUV.

The aim of this research is to develop a novel robotics vision system at conceptual level, in order to assist AUV's interpretation of underwater oceanic scenes for the purpose of object tracking and intelligent navigation. Underwater images captured containing object of interest (Pipeline), simulated seabed, water and other unwanted noises. Image processing techniques i.e. morphological filtering, noise removal, edge detection, etc, are performed on the images in order to extract subjective uncertainties of the object of interest. Subjective uncertainties became multiple input of a fuzzy inference system. Fuzzy rules and membership function is determined in this project. The fuzzy output is a crisp value of the direction for navigation or decision on the control action.

*2.1 Image Processing Operations*

For this vision system, image analysis is the first process and the end product shall be the extraction of high-level information for computer analysis and manipulation. This high-level information is actually the morphological parameter for the input of a fuzzy inferences system (linguistic representation of terrain features).

When an RGB image is loaded, it is converted into gray scale image. RGB image as shown in Figure 1. Then gray-level thresholding is performed to extract the object or region of interest from the background. The intensity levels of the object of interest are identified. The binary image B[i,j], is obtained using object of interest's intensity values in the range of [$T_1$, $T_2$] for the original gray image F[i,j]. That is,

$$B[i,j] = \begin{cases} 1 & T_1 < F[i,j] \leq T_2 \\ 0 & otherwise \end{cases} \quad (1)$$

The thresholding process producing a binary image with a large region of connected pixels (object of interest) and large amount of small region of connected pixels (noise). Each region is labeled and the largest connected region is identified as object of interest. In the labeling process, the connected pixels are labeled as either object of interest or unwanted objects by examining their connectivity's (eight-connectivity) to neighboring pixels. Label will be assigned to the largest connected region that represents the object of interest.

At this stage, feature extraction is considered completed. Object of interest is actually pipeline laid along the perspective view of the camera. Image is then horizontally, from image bottom to top, divided into 5 segment and be processed separately for terrain features as multiple steps of inputs for the fuzzy controller. In order to investigate more closely each specific area within the image segment, each segment is further divided into 6 predefined sub segments in the image. Each sub segment (as illustrated by Figure 2) is defined as follows.

- Sub segment 1 = Upper left segment of the image
- Sub segment 2 = Upper right segment of the image
- Sub segment 3 = Lower left segment of the image
- Sub segment 4 = Lower right segment of the image
- Sub segment 5 = Upper segment of the image
- Sub segment 6 = Lower segment of the image.

A mask image with constant intensity is then laid on the image as shown in Figure 3. This is actually an image addition process whereby it will produce a lighter (highest intensity value) area when intersects the region of interest. The remaining region with highest intensity value then be calculated its coverage area in the image as shown in Figure 4. The area, A of the image is determined by.

$$A = \sum_{i=1}^{n} \sum_{j=1}^{m} B[i,j] \quad (2)$$

Sub segment 5-6 are being determined its location relative to the image center. Coverage area and location of object of interest in each sub segment is finally be accumulated as multiple input of the fuzzy inference system.

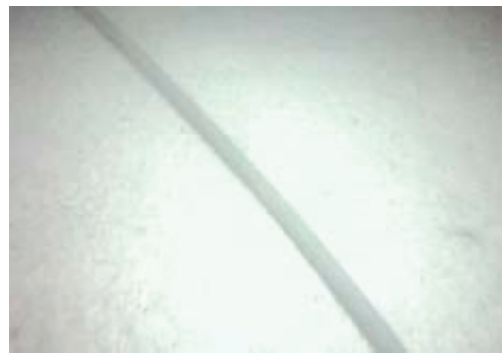

Fig. 1. Typical input image (RGB)

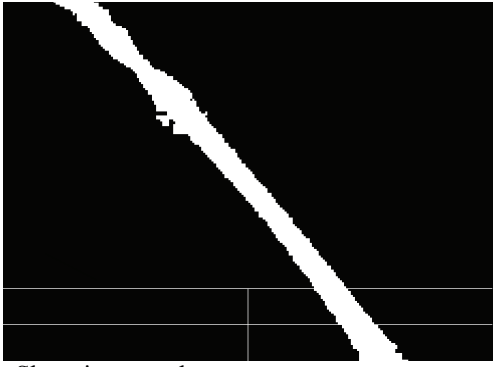
Fig. 2. Show image sub segment

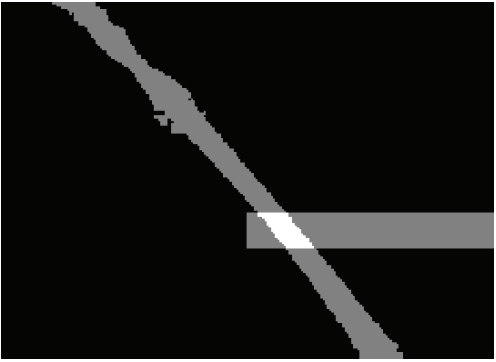
Fig. 3. Mask on threshold, removed noise image

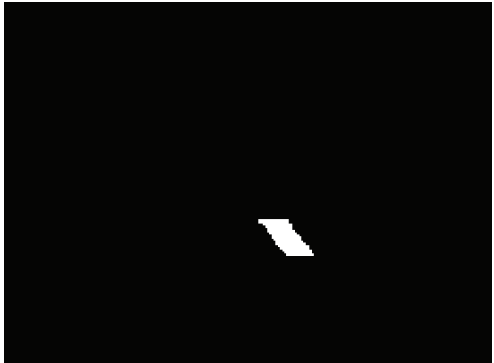
Fig. 4. Acquired area information

*2.1 The Fuzzy Inference System (FIS)*

The fuzzy controller is designed to automate how a human expert who is successful at this task would control the system. The multiple inputs to the controller are variables defining the state of the camera with respect to the pipeline, and the single output is the steering command set point. Consider the situation illustrated by figure 5. The fuzzy logic is used to interpret this heuristic in order to generate the steering command set point. In this case, the set point of AUV has a certain amount ($\Delta X$) to the right.

Basically, a human operator does not require a crisp / accurate visual input for mission completion. There are total of six inputs based on the image processing algorithm.

- Input variable 1, $x_1$ = *Pipeline area at upper left segment in the image*
  Input variable 1 fuzzy term set, $T(x_1)$ = {*Small, Medium, Large*}
  Input variable 1 universe of discourse, $U(x_1)$ = [0.1 - 1.0]
- Input variable 2, $x_2$ = *Pipeline area at upper right segment in the image*
  Input variable 2 fuzzy term set, $T(x_2)$ = {*Small, Medium, Large*}
  Input variable 2 universe of discourse, $U(x_2)$ = [0.1 - 1.0]
- Input variable 3, $x_3$ = *Pipeline area at lower left segment in the image*
  Input variable 3 fuzzy term set, $T(x_3)$ = {*Small, Medium, Large*}
  Input variable 3 universe of discourse, $U(x_3)$ = [0.1 - 1.0]
- Input variable 4, $x_4$ = *Pipeline area at lower right segment in the image*
  Input variable 4 fuzzy term set, $T(x_4)$ = {*Small, Medium, Large*}
  Input variable 4 universe of discourse, $U(x_4)$ = [0.1 - 1.0]
- Input variable 5, $x_5$ = *End point of pipeline relative to image center point*
  Input variable 5 fuzzy term set, $T(x_5)$ = {*Left, Center, Right*}
  Input variable 5 universe of discourse, $U(x_5)$ = [0.1 - 1.0]
- Input variable 6, $x_6$ = *Beginning point of pipeline relative to image center point*
  Input variable 6 fuzzy term set, $T(x_6)$ = {*Left, Center, Right*}
  Input variable 6 universe of discourse, $U(x_6)$ = [0.1 - 1.0]

The only fuzzy output.

- Output variable 1, $y_1$ = *AUV steering command set point*
  Output variable 1 fuzzy term set, $T(y_1)$ = {*Turn left, Go straight, Turn right*}
  Output variable 1 universe of discourse, $V(y_1)$ = [0 - 180]

The input vector, x is.

$$x = (x_1, x_2, x_3, x_4, x_5, x_6)^T \qquad (3)$$

The output vector, y is.

$$y = (y_1)^T \qquad (4)$$

Gaussian and π-shaped membership functions are selected in this case to map the input to the output. Gaussian curves depend on two parameters σ and c and are represented by.

$$f(x; \sigma, c) = \exp\left[\frac{-(x-c)^2}{2\sigma^2}\right] \qquad (5)$$

π-shaped membership function are represented by.

$$f(x;b,c) = \begin{cases} S(x;c-b, c-b/2, c) & \text{for } x \le c \\ 1 - s(x;c, c+b/2, c+b) & \text{for } x > c \end{cases} \quad (6)$$

where $S(x; a, b, c)$ represents a membership function defined as.

$$S(x;a,b,c) = \begin{cases} 0 & \text{for } x < a \\ \dfrac{2(x-a)^2}{(c-a)^2} & \text{for } a \le x < b \\ 1 - \dfrac{2(x-a)^2}{(c-a)^2} & \text{for } b \le x \le c \\ 1 & \text{for } x > c \end{cases} \quad (7)$$

In the above equation, σ, a, b and c are the parameters that are adjusted to fit the desired membership data. Typical input variable and output variable membership function plot are shown in figure 6 and figure 7.

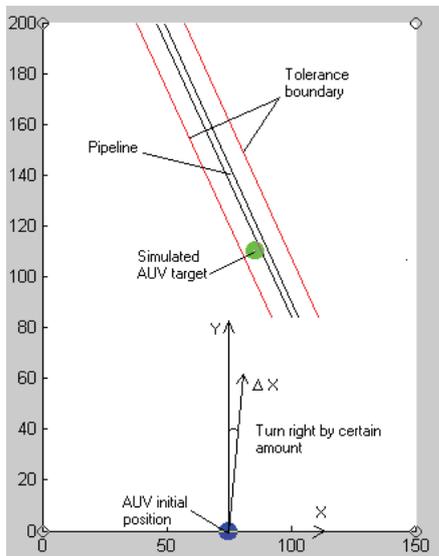
Fig. 5. Illustration of tracking strategy

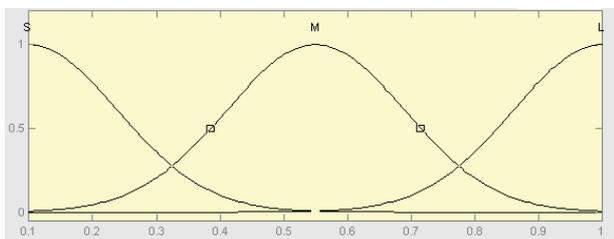
Fig. 6. Typical input variable membership function plot

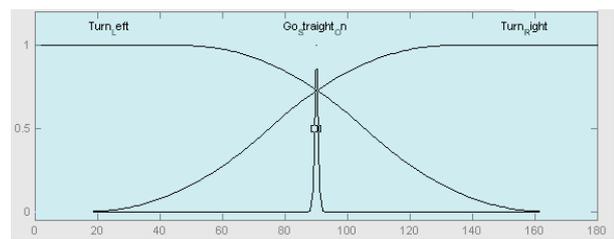
Fig. 7. Typical output variable membership function plot

There are totally 13 fuzzy control rules. The rule base as shown in figure 8.

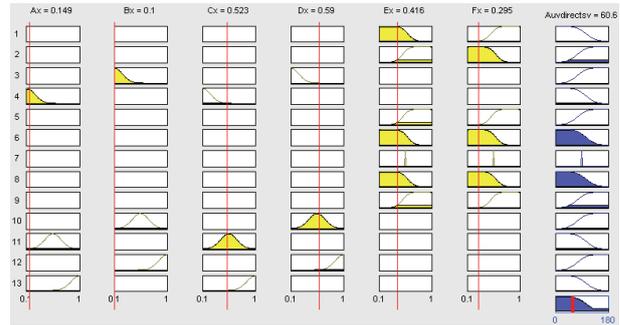
Fig. 8. Rule viewer for fuzzy controller

In order to obtain a crisp output, the output fuzzy set is then aggregated and fed into a centroid (center of gravity) method defuzzification process. The defuzzifier determines the actual actuating signal, $y'$ as follows.

$$y' = \dfrac{\sum_{i=1}^{13} y_i \mu_B(y_i)}{\sum_{i=1}^{13} \mu_B(y_i)} \quad (8)$$

## 3. Simulation and Experimental Results

The simulation procedure is as follows:
a. Defined the envelope curve (working area) of prototype.
b. Given the real position and orientation of pipeline defined on a grid of coordinates.
c. Predefined the AUV drift tolerance limit (±8.0cm) away from the actual pipeline location.
d. Initiate the algorithm.
e. AUV navigating paths are recorded and visualized graphically.

The algorithm has been tested on computer and prototype simulations. For comparative purposes, the results before and after fuzzy tuning are presented. Typical examples of results before fuzzy tuning are shown in Figure 9 and Table 1.

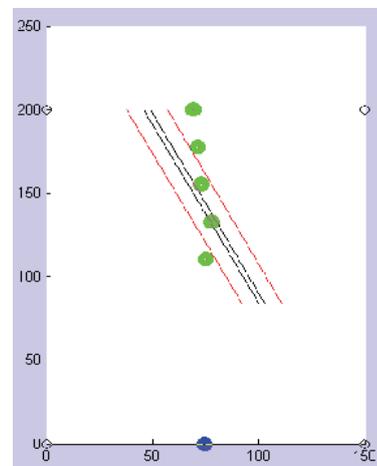
Fig. 9. AUV path (no proper tuning)

| AUV path | Actual location x-axis (cm) | Simulated result x-axis (cm) | Drift (cm) | Percentage of Drift (%) |
|---|---|---|---|---|
| 5 | 47.5 | 69.5 | +22.0 | 275.0 |
| 4 | 58.5 | 71.7 | +13.2 | 165.0 |
| 3 | 69.6 | 73.3 | +3.7 | 46.3 |
| 2 | 80.8 | 78.3 | -2.5 | 31.3 |
| 1 | 91.9 | 75.7 | -16.2 | 202.5 |

Table 1. Data recorded (without proper tuning)

Typical examples of results after fuzzy tuning are shown in Figure 10 and Table 2.

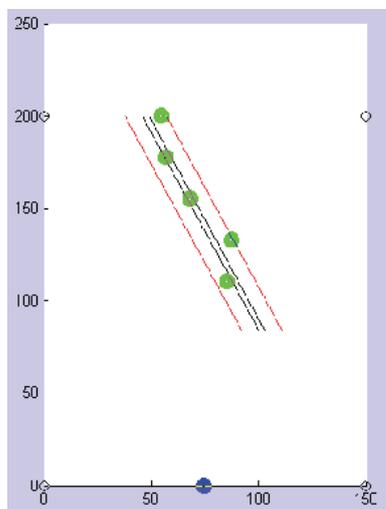

Fig. 10. AUV path (with proper tuning)

| AUV path | Actual location x-axis (cm) | Simulated result x-axis (cm) | Drift (cm) | Percentage of Drift (%) |
|---|---|---|---|---|
| 5 | 47.5 | 55.2 | +7.7 | 96.3 |
| 4 | 58.5 | 57.4 | -1.1 | 13.8 |
| 3 | 69.6 | 68.5 | -1.1 | 13.8 |
| 2 | 80.8 | 88.1 | +7.3 | 91.3 |
| 1 | 91.9 | 85.9 | -6.0 | 75.0 |

Table 2. Data recorded (with proper tuning)

The simulation results show that the drift within tolerance limit is achievable when proper tuning (training) is applied to the fuzzy system. The percentage of drift is considered acceptable , as long as it is less than 100%, since this implies the path is within the boundary. The effectiveness of the system has been further demonstrated with different target orientation and lighting condition.

## 4. Conclusions

This paper has introduced a new technique for AUV target tracking and navigation. The image processing algorithm developed is capable of extracting qualitative information of the terrain required by human operators to maneuver ROV for pipeline tracking. It is interesting to note that fuzzy control system developed is able to mimic human operators' inherent ability for deciding on acceptable control actions. This has been verified experimentally and the result is favourable that is within 8.0 cm of drift tolerance limit in a 1.5m x 2.0m working envelope. One of the most interesting parts being the system ability to perform target tracking and navigation from the knowledge of interpreting image grabbed in perspective view from the terrain.

It should also be noted that the system offer another human-like method of representing human experience and knowledge of operating a ROV, rather than being expressed in differential equations in the common PID-controller. Obviously, the system does not require sophisticated image processing algorithm such as Kalman filtering or Hough transform techniques. All input variable required are merely an approximate value for mission completion, just like a human vision system. The simplicity of the system is further recognized when a priori knowledge of the terrain is not necessary as part of the algorithm. Currently a priori knowledge is required by some of the available pipeline tracking techniques such as (Evans, J, et al, 2003) and (Arjuna Balasuriya and Ura, T, 2002). The processing time is therefore reduced.

In general the whole computational process for this prototype is complex and it usually takes about 60 seconds to arrive at its desired output for 5 steps (22.5cm for each step), which is not practical for commercial standard requirement that is at least 4 knot (2m/s) of AUV speed. Commercial standard requirement of a survey AUV can be found in (Bingham, D. ,2002). However, the proposed system would be a workable concept for its capability to look forward and perceive the terrain from perspective view.

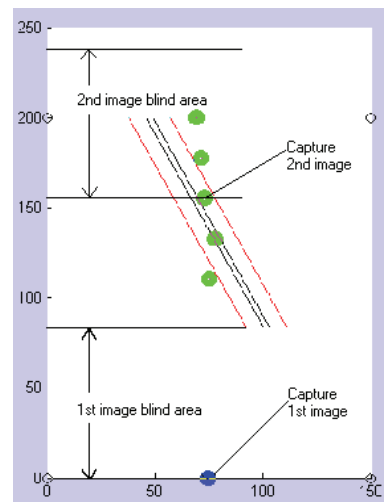

Fig. 11. AUV path and its image capturing procedure

As illustrated in Figure 11, the perceived conditions from the second image captured could be processed concurrently while the AUV completing the forth and fifth step based on the previous image information. This will improve the processing time to support high speed AUV application.

In addition, further studies on improving the program structure and calculation steps may help to achieve better computation time. Future development of transputer for parallel processing or higher speed processor can also be expected to bring the system into practical use.